\documentclass[runningheads]{llncs}

\usepackage[T1]{fontenc}
\usepackage{graphicx}
\newcommand*\samethanks[1][\value{footnote}]{\footnotemark[#1]}

\usepackage{amsmath,amssymb} 

\usepackage{float}
\usepackage{siunitx}
\usepackage{newclude}

\usepackage{multirow}

\usepackage{wrapfig}
\usepackage{tabularray}

\usepackage{numprint}

\usepackage{placeins}

\usepackage[numbers,sort&compress]{natbib}
\usepackage{hyperref}
\usepackage{cleveref}

\begin{document}
\title{\emph{blob loss:} instance imbalance aware loss functions for semantic segmentation}

%
%

\author{Florian Kofler\inst{1,2,3} \and
Suprosanna Shit\inst{1,2} \and
Ivan Ezhov\inst{1,2} \and
Lucas Fidon\inst{4} \and
\\ Izabela Horvath\inst{1,5} \and
Rami Al-Maskari\inst{1,5} \and
Hongwei Li\inst{1,13} \and
Harsharan Bhatia\inst{5,6} \and
Timo Loehr\inst{1,3} \and
Marie Piraud\inst{7} \and
Ali Erturk\inst{5,6,8,9} \and
\\ Jan Kirschke\inst{3} \and
Jan C. Peeken\inst{10,11,12} \and
Tom Vercauteren\inst{4} \and
Claus Zimmer\inst{3} \and
\\ Benedikt Wiestler\inst{3} \thanks{contributed equally as senior authors} \and
Bjoern Menze\inst{1,13 \samethanks}
}

\authorrunning{F. Kofler et al.}
%

\institute{
Department of Informatics, Technical University Munich, Germany \and
TranslaTUM - Central Institute for Translational Cancer Research, Technical University of Munich, Germany \and
Department of Diagnostic and Interventional Neuroradiology, School of Medicine, Klinikum rechts der Isar, Technical University of Munich, Germany \and
School of Biomedical Engineering \& Imaging Sciences, King's College London, United Kingdom \and
Institute for Tissue Engineering and Regenerative Medicine, Helmholtz Institute Munich (iTERM), Germany \and
Institute for Stroke and dementia research (ISD), University Hospital, LMU Munich, Germany \and
Helmholtz AI, Helmholtz Munich, Germany \and
Graduate school of neuroscience (GSN), Munich, Germany \and
Munich cluster for systems neurology (Synergy), Munich, Germany \and
Department of Radiation Oncology, Klinikum rechts der Isar, Technical University of Munich, Germany \and
Institute of Radiation Medicine (IRM), Department of Radiation Sciences (DRS), Helmholtz Zentrum, Munich, Germany \and
Deutsches Konsortium für Translationale Krebsforschung (DKTK), Partner Site Munich, Germany \and
Department of Quantitative Biomedicine, University of Zurich, Switzerland
}






\maketitle              
%


\clearpage
\begin{abstract}
Deep convolutional neural networks (CNN) have proven to be remarkably effective in semantic segmentation tasks.
Most popular loss functions were introduced targeting improved volumetric scores, such as the Dice coefficient (DSC).
By design, DSC can tackle class imbalance, however, it does not recognize instance imbalance within a class.
As a result, a large foreground instance can dominate minor instances and still produce a satisfactory DSC.
Nevertheless, detecting tiny instances is crucial for many applications, such as disease monitoring.
For example, it is imperative to locate and surveil small-scale lesions in the follow-up of multiple sclerosis patients.
We propose a novel family of loss functions, \emph{blob loss}, primarily aimed at maximizing instance-level detection metrics, such as \emph{F1} score and \emph{sensitivity}.
\emph{Blob loss} is designed for semantic segmentation problems where detecting multiple instances matters.
We extensively evaluate a DSC-based \emph{blob loss} in five complex 3D semantic segmentation tasks featuring pronounced instance heterogeneity in terms of texture and morphology.
Compared to soft Dice loss, we achieve \emph{5\%} improvement for MS lesions, \emph{3\%} improvement for liver tumor, and an average \emph{2\%} improvement for microscopy segmentation tasks considering \emph{F1} score.
\end{abstract}

\section{Introduction}
In recent years convolutional neural networks (CNN) have gained increasing popularity for complex machine learning tasks, such as \emph{semantic segmentation}.
In \emph{semantic segmentation}, one segments object from different classes without differentiating multiple instances within a single class.
In contrast, \emph{instance segmentation} explicitly takes multiple instances into account, which involves simultaneous localization and segmentation.
While U-net variants \cite{ronneberger2015u} still represent the state-of-the-art to address semantic segmentation, \emph{Mask-RCNN} and its variants dominate \emph{instance segmentation} \cite{he2017mask}.
The scarcity of training data often hinders the application of back-bone-dependent Mask RCNNs, while U-Nets have proven to be less data-hungry  \cite{caicedo2019nucleus}.


However, many semantic segmentation tasks feature relevant instance imbalance, where large instances dominate over smaller ones within a class, as illustrated in \Cref{fig:problem_statement}.
Instances can vary not only with regard to size but also texture and other morphological features.
U-nets trained with existing loss functions, such as Soft Dice \cite{milletari2016v,salehi2017tversky,ma2021loss,eelbode2020optimization,sudre2017generalised}, cannot address this.
Instance imbalance is particularly pronounced and significant in medical applications:
For example, even a single new multiple sclerosis (MS) lesion can impact the therapy decision.
Despite many ways to compensate for class-imbalance \cite{sudre2017generalised,fidon2017generalised,berman2018lovasz,rahman2016optimizing}, there is a notable void in addressing instance imbalance in semantic segmentation settings.
Additionally, established metrics have been shown to correlate insufficiently with expert assessment \cite{kofler2021using}.

\noindent\textbf{Contribution:}
We propose \emph{blob loss}, a novel framework to equip semantic segmentation models with instance imbalance awareness.
This is achieved by dedicating a specific loss term to each instance without the necessity of instance-wise prediction.
\emph{Blob loss} represents a method to convert any loss function into a novel instance imbalance aware loss function for semantic segmentation problems designed to optimize detection metrics.
We evaluate its performance on five complex three-dimensional (3D)  semantic segmentation tasks, for which the discovery of miniature structures matters.
We demonstrate that extending soft Dice loss to a \emph{blob loss} improves detection performance in these multi-instance semantic segmentation tasks significantly.
Furthermore, we also achieve volumetric improvements in some cases.

\noindent\textbf{Related work:} \citet{sirinukunwattana2015stochastic} suggested an instance-based Dice metric for evaluating segmentation performance.
\citet{salehi2017tversky} were among the first to propose a loss function, called \emph{Tversky loss}, for semantic segmentation of multiple sclerosis lesions in magnetic resonance imaging (MR), trying to improve detection metrics.
Similarly, \citet{zhu2019anatomynet} introduced Focal Loss, initially designed for object detection tasks \cite{lin2017focal}, into medical semantic segmentation tasks.

There have been few recent attempts aiming for a solution to instance imbalance.
\citet{zhang2021all} propose an auxiliary lesion-level sphere prediction task.
However, they do not explicitly consider each instance separately.
\citet{shirokikh2020universal} propose an instance-weighted loss function where a global weight map is inversely proportional to the size of the instances.
However, unlike size, not all types of imbalance, such as morphology or texture, can be quantified easily, limiting the method's applicability.

\clearpage
\section{Methods}
First, we introduce the problem of instance imbalance in semantic segmentation tasks.
Then we present our proposed \emph{blob loss} functions.



\begin{figure}[H]
    \centering
    \includegraphics[width=1.0\textwidth]{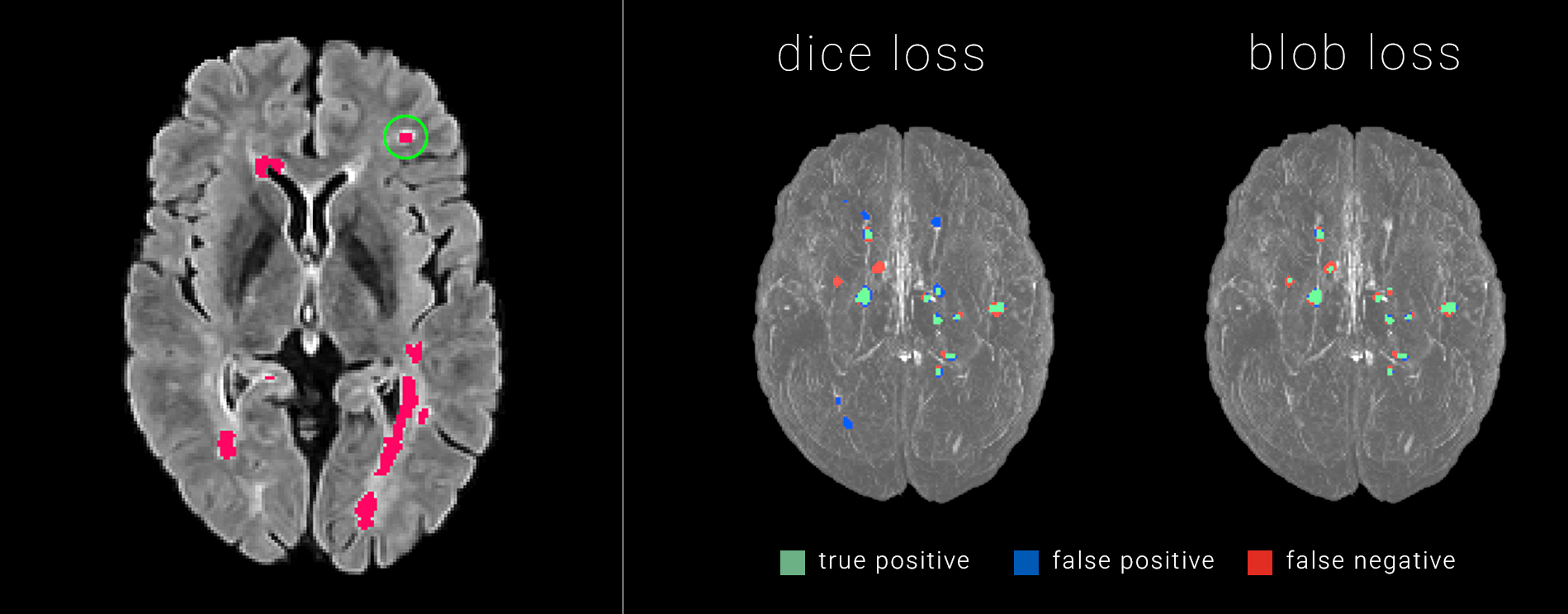}
    \caption{
    \emph{Problem statement (left):}
    The Dice coefficient (DSC) for the segmentation with vs. without a lesion, encircled in green, is: \emph{0.9806}.
    Therefore, the segmentations are hardly distinguishable in terms of DSC.
    However, from a clinical perspective, the difference is important as the detection of a single lesion can affect treatment decisions. \\
    \emph{Comparison of segmentation performance (right)}:
    Maximum intensity projections of the FLAIR images overlayed with segmentations for \emph{dice} and \emph{blob dice}.
    Lesions are colored according to their detection status:
    Green for \emph{true positive};
    Blue for \emph{false positive};
    Red for \emph{false negative}.
    For this particular patient, applying the transformation to a \emph{blob loss} improves \emph{F1} from \emph{0.74} to \emph{1.0} and the volumetric Dice coefficient from \emph{0.56} to \emph{0.70} and the latter is caused by an increase in \emph{volumetric precision} from \emph{0.48} to \emph{0.75}, while the 
    \emph{volumetric sensitivity} remains constant at \emph{0.66}.
    }
    \label{fig:problem_statement}
\end{figure}

\noindent\textbf{Problem statement:}
Large foreground areas dominate the calculation of established volumetric metrics (or losses); see \Cref{fig:problem_statement}.
%

This is because the volumetric measures only accumulate true or false predictions on a voxel level but not at the instance level.
Therefore, training models with volumetry-based loss functions, such as soft Dice loss (\emph{dice}), often leads to unsatisfactory instance detection performance.
To achieve a better instance detection performance, it is necessary to take instance imbalance into account.
Instance imbalance can be of many categories, such as morphology and texture.
Importantly, instance imbalance often cannot be easily specified and quantified for use in CNN training, for example, as instance weights in the loss function.
Thus, using conventional methods, it is difficult to incorporate instance imbalance in CNN training.
Our objective is to design loss functions to compensate for the instance imbalance while being agnostic to the instance imbalance type.
Therefore, we aim to dissect the image domain in an instance-wise fashion:

\noindent\textbf{\emph{blob loss} formulation:}
Consider a generic volumetric loss function $\mathcal{L}$ and image domain $\Omega$ and foreground domain $P$.
Formally our objective is to find an instance-specific subdomain $\Omega_{n}\subseteq\Omega$ corresponding to the $n^{th}$ instance such that $\mathcal{L}$ acting on $\Omega_{n}$ is aware of instance imbalance.
The criteria to obtain these subsets $\{\Omega_{n}\}_{n=1}^{N}$ are such that $\Omega_{i}\cap\Omega_{j}\cap P = \phi; \forall (i,j),~s.t.~ 1\leq i,j \leq N, i\neq j$ and $\cup_{n=1}^{N} \Omega_{n}= \Omega$.
In simple terms, the subsets $\{\Omega_{n}\}_{n=1}^{N}$ need to be mutually exclusive regarding foreground and collectively exhaustive with regard to the whole image domain.

\label{sec:definition}
To formalize \emph{blob loss}, we address instance imbalance within a binary semantic segmentation framework.
At the same time, we remain agnostic towards particular instance attributes and do not incorporate these in the loss function.
To this extent, we propose to leverage the existing reference annotations and formally propose a novel family of instance-aware loss functions.

\begin{figure}
    \centering    \includegraphics[width=\linewidth]{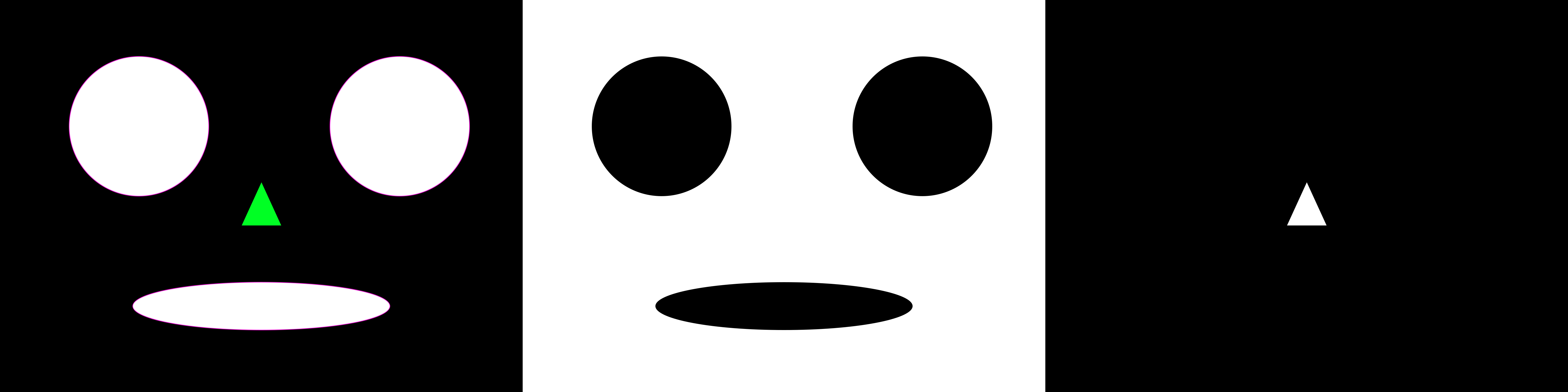}
    \caption{
    Masking process described in \Cref{eq:masking}.
    \emph{Left:} the global ground truth label (GT), with the  $n^{th}$ instance highlighted in green.
    \emph{Middle:} The loss mask $\Omega_n$ for the $n^{th}$ instance (MASK) for multiplication with the network outputs.
    \emph{Right:} the label used for the computation of the local \emph{blob loss} for the $n^{th}$ instance.
    This process is repeated for every instance.
    }
    \label{fig:masking}
\end{figure}

Consider a segmentation problem with $N$ instances; for different input images, $N$ can vary from few to many.
Specifically, we propose to compute the instance-specific domain $\Omega_{n}$ by excluding all but the $n^{th}$ foreground from the whole image domain $\Omega$, see \Cref{eq:omegan}:
\begin{equation}
    \Omega_n = \Omega \setminus \cup_{j=1,\,j\neq n}^N P_j
        \label{eq:omegan}
\end{equation}

where $P_j$ is the foreground domain for $j^{th}$ instances of $P$.
This masking process is illustrated by \Cref{fig:masking}.
It is worth noting that the background voxels are included in every $\Omega_n$.

We propose to convert any loss function $\mathcal{L}$ for binary semantic segmentation into an instance-aware loss function $\mathcal{L}_{blob}$ defined as:
\begin{equation}
    \mathcal{L}_{blob}\left((p_i)_{i\in \Omega}, (g_i)_{i\in \Omega}\right)
    = \frac{1}{N}
    \sum_{n=1}^{N} \mathcal{L}
    \left(
    (p_i)_{i\in \Omega_n}, (g_i)_{i\in \Omega_n}
    \right)
    \label{eq:masking}
\end{equation}

where $\{g_i\}_{i\in \Omega}$ is the ground-truth segmentation,
$\{p_i\}_{i\in \Omega}$ is the predicted segmentation, 
$N$ is the number of instances in the foreground.

As our goal is to assign equal importance to all instances irrespective of their size, shape, texture, and other topological attributes, we average over all instances.

To compute the total loss for a volume, we combine the instance-wise Loss component from \Cref{eq:masking} with a global component to obtain the final Loss:
\begin{equation}
\mathcal{L}_{total}  = \alpha \mathcal{L}_{global} + \beta \mathcal{L}_{blob}
\label{eq:whole}
\end{equation}

where $\alpha$ and $\beta$ denote the weights for the global and instance constraint $\mathcal{L}_{blob}$.
We (anonymously) provide a sample Pytorch implementation of a \emph{dice}-based \emph{blob loss} on \href{https://pastebin.com/Nqy5VyHR}{GitHub}.
In order to accelerate our training, we precompute the instances, here defined as connected components using \emph{cc3d} \cite{william_silversmith_2021_5535251}, version \emph{3.2.1}.


\noindent\textbf{Model training:}
\label{sec:cnn_training}
For all our experiments, we use a basic 3D U-Net implemented via \href{https://docs.monai.io/en/latest/networks.html#basicunet}{MONAI} inspired by \cite{falk2019u} and further depicted in supplementary materials.
Furthermore, we use a dropout ratio of \emph{0.1} and employ \emph{mish} as activation function \cite{misra2019mish}. Otherwise, we stick to the default parameters of the U-Net implementation.

\noindent\textbf{Loss functions for comparison:}
As baselines we use the MONAI implementations of soft Dice loss (\emph{dice}) and Tversky loss (\emph{tversky}) \cite{salehi2017tversky}.
For \emph{tversky}, we always use the standard parameters of \emph{$\alpha$ = 0.3} and \emph{$\beta$ = 0.7} suggested by the authors in the original publication \cite{salehi2017tversky}.
For comparison we create \emph{blob dice}, by transforming the standard \emph{dice} into a \emph{blob loss} using our conversion method \Cref{eq:masking}.
The final loss is obtained by employing \emph{dice} in the  $\mathcal{L}_{global}$ and $\mathcal{L}_{blob}$ terms of the proposed total loss \Cref{eq:whole}.
In analog fashion, we derive \emph{blob tversky}.
Furthermore, we compare against \emph{inverse weighting (iw)}, the globally weighted loss function of \citet{shirokikh2020universal}.
For this, we use the official \href{https://github.com/neuro-ml/inverse_weighting}{GitHub implementation} to compute the weight maps and loss and deploy these in our training pipelines.

\noindent\textbf{Training procedure:}
Our CNNs are trained on multiple cuboid-shaped crops per batch element, with higher resolution in the axial dimension, enabling the learning of contextual image features.
The crops are randomly sampled around a center voxel that consists of \emph{foreground} with a \emph{95\%} probability.
We consider one epoch as one full iteration of forward and backward passes through all batches of the training set.
For all training, \emph{Ranger21} \cite{wright2021ranger21} serves as our optimizer.
For each experiment, we keep the initial learning rate (lr) constant between training runs.
Depending on the segmentation task, we deploy varying suitable image normalization strategies.
For comparability, we keep all training parameters except for the loss functions constant on a segmentation task basis and stick to this standard training procedure.

\noindent\textbf{Training-test split and model selection:}
Given the high heterogeneity of our bio-medical datasets and the limited availability of high-quality ground truth annotations due to the very costly labeling procedures requiring domain experts, we do not set aside data for validation and therefore do not conduct model selection.
Instead, inspired by \cite{isensee2019nnu}, we split our data \emph{80:20} into training and test set and evaluate on the last checkpoint of the model training.
As an exception, the MS dataset comes with predefined training, validation, and test set splits;
therefore, we additionally evaluate the \emph{best} model checkpoint, meaning the model with the lowest loss on the validation set.
As we are more interested in \emph{blob loss'} generalization capabilities than exact quantification of improvements on particular datasets, we prioritize a broad validation on multiple datasets over cross-validation.


\noindent\textbf{Technical details:}
Our experiments were conducted using NVIDIA RTX8000, RTX6000, RTX3090, and A6000 GPUs using CUDA version  \emph{11.4} in conjunction with Pytorch version \emph{1.9.1} and MONAI version \emph{0.7.0}.

\subsection{Evaluation Metrics and Interpretation}
\label{sec:interpretation}

\noindent\textbf{Metrics:} We obtain global, volumetric performance measures from \emph{pymia} \cite{jungo2021pymia}.
In addition to DSC, we also evaluate \emph{volumetric sensitivity (S)}, \emph{volumetric precision (P)}, and the \emph{Surface Dice similarity coefficient (SDSC)}.
To compute instance-wise detection metrics, namely instance F1 (\emph{F1}), \emph{instance sensitivity (IS)} and \emph{instance precision (IP)},  we employ a proven evaluation pipeline from \citet{pan2019deep}.

\begin{wrapfigure}{r}{0.5\textwidth}
    \centering
    \vspace{-8mm}
    \includegraphics[width=1.0\linewidth]{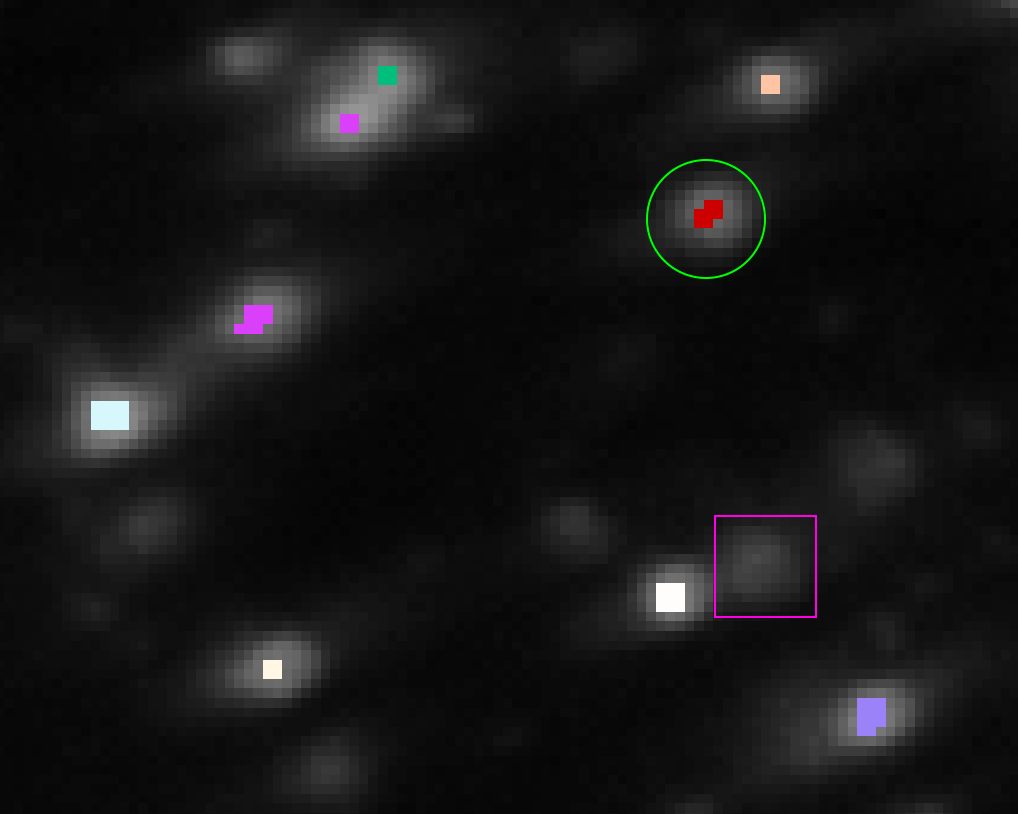}
    \vspace{-5mm}
    \caption{
    Zoomed in 2D view on a volume of the SHANEL \cite{zhao2020cellular} dataset.
    The overlayed labels are colored according to a 3D connected component analysis.
    The expert biologists did not label each foreground object in every slice, e.g., the magenta-colored square.
    Furthermore, the contours of the structures are imperfectly segmented, for instance, the red label within the bright green circle.
    These effects can partially be attributed to the ambiguity of the light-sheet microscopy signal \cite{kofler2022approaching}.
    However, they are also observed in the human annotations of the MS and LiTS dataset.
    }
    \label{fig:label_quality}
\vspace{-10mm}
\end{wrapfigure}

\noindent\textbf{Interpretation:}
By design, human annotators tend to overlook tiny structures.
For comparison, human annotators initially missed \emph{29\%} of micrometastases when labeling the DeepMACT light-sheet microscopy dataset \cite{pan2019deep}.
Therefore, the likelihood of a structure being correctly labeled in the ground truth is much higher for foreground than for background structures.
Additionally, human annotators have a tendency to label a structure's center but do not perfectly trace its contours.
Both phenomena are illustrated in \Cref{fig:label_quality}.
These effects are particularly pronounced for microscopy datasets, which often feature thousands of blobs.
These factors are important to keep in mind when interpreting the results.
Consequently, volumetric - and instance sensitivity are much more informative than volumetric and instance precision.

\clearpage
\section{Experiments}

To validate \emph{blob loss}, we train segmentation models on a selection of datasets from different 3D imaging modalities, namely brain MR, thorax CT, and light-sheet microscopy.
We select datasets featuring a variety of fragmented semantic segmentation problems.
\Cref{fig:datasets} features an overview over blob counts, volumetrics, and shape features in the datasets.
For simplicity, we use the default values $\alpha = 2.0$ and $\beta = 1.0$ across all experiments.

\noindent\textbf{\emph{Multiple Sclerosis (MRI):}}
The Multiple Sclerosis (MS) dataset, comprising 521 single timepoint MRI examinations of patients with MS, was collected for internal validation of MS lesion segmentation algorithms.
The patients come from a representative, institutional cohort covering all stages (in terms of time from disease onset) and forms (relapsing-remitting, progressive) of MS.
A 3D T1w and a 3D FLAIR sequence were acquired on a 3 Tesla \emph{Philips Achieva} scanner.
All 3D volumes feature \emph{193x193x229} voxels in 1mm isotropic resolution.
The dataset divides into a fixed training set of 200, a validation set of 21, and a test set of 200 cases.
The annotations feature a total of \emph{4791} blobs, with \emph{25.69$\pm$23.01} blobs per sample.
Expert neuroradiologists annotated the MS lesions manually and ensured pristine ground truth quality with consensus voting.

For all training runs of \emph{500} epochs, we set the initial learning rate to \emph{1e-2} and the batch size to \emph{4}.
The networks are trained on a single GPU using \emph{2} random crops with a patch size of \emph{192x192x32} voxels per batch element after applying a \emph{min/max} normalization.
As the MS dataset comes with a predefined validation set of 21 images, we also save the checkpoint with the lowest loss on the validation set and compare it to the respective last checkpoint of the training.
In addition to the standard \emph{dice}, we also compare against \emph{tversky}.
Furthermore, we conduct an ablation study to find out how the performance metrics are affected by choosing different values for $\alpha$ and $\beta$.

\noindent\textbf{Liver Tumors - LiTS (CT):}
To develop an understanding of \emph{blob loss} performance on other imaging modalities, we train a model for segmenting liver tumors on CT images of the \emph{LiTS} challenge \cite{bilic2019liver}.
The dataset consists of varying high-resolution CT images of the abdomen.
The challenge's original task was segmenting liver and liver tumor tissue.
As we are primarily interested in segmenting small fragmented structures, we limit our experiments to the liver area and segment only liver tumor tissue (in contrast to tumors, the liver represents a huge solid structure, and we are interested in blobs).
We split the publicly available training set into 104 images for training and 27 for testing.
The annotations were created by expert radiologists and feature a total of \emph{908} blobs, with \emph{12.39$\pm$14.92} blobs per sample.

For all training runs of \emph{500} epochs, we set the initial learning rate to \emph{1e-2} and the batch size to \emph{2}. 
The networks are trained on two GPUs in parallel using \emph{2} random crops with a patch size of \emph{192x192x64} voxels per batch element.
We apply normalization based on windowing on the Hounsfield (HU) scale.
Therefore, we define a normalization window suitable for liver tumor segmentation around center \emph{30 HU} with a width of \emph{150 HU}, and \emph{20\%} added tolerance.

\noindent\textbf{\emph{DISCO-MS (light-sheet microscopy)}}
To develop an understanding for \emph{blob loss} performance on other imaging modalities, we train a model for segmenting Amyloid plaques in light-sheet microscopy images of the \emph{DISCO-MS} dataset \cite{bhatia2021proteomics}.

The volumes of \emph{300x300x300} voxels resolution contain cleared tissue of mouse brain.
We split the publicly available dataset into 41 volumes for training and six for testing.
The annotations feature a total of \emph{988} blobs, with \emph{28.32$\pm$24.44} blobs per sample.
Even though the label quality is very high, the results should still be interpreted with care following the guidelines in \Cref{sec:interpretation}.

For all training runs of \emph{800} epochs, we set the initial learning rate to \emph{1e-3} and the batch size to \emph{6}.
As our initial model trained with \emph{dice} does not produce satisfactory results, we furthermore try learning rates of \emph{1e-2}, \emph{3e-4} and \emph{1e-4}, following the heuristics suggested by \cite{bengio2012practical} without success.
The networks are trained on two GPUs in parallel using \emph{2} random crops with a patch size of \emph{192x192x64} per batch element.
The images are globally normalized, using a minimum and maximum threshold defined by the \emph{0.5} and \emph{99.5} percentile.

\noindent\textbf{\emph{SHANEL (light-sheet microscopy)}}
For further validation, we evaluate neuron segmentation in light-sheet microscopy images of the \emph{SHANEL} dataset \cite{zhao2020cellular}.
The volumes of \emph{200x200x200} voxels resolution contain cleared human brain tissue from the primary visual cortex, the primary sensory cortex, the primary motor cortex, and the hippocampus. 
We split this publicly available dataset into nine volumes for training and three for testing.
The annotations feature a total of \emph{20684} blobs, with \emph{992.14$\pm$689.39} blobs per sample.
As the data is more sparsely annotated than DISCO-MS, F1 and especially DSC should be interpreted with great care, as described in \Cref{sec:interpretation}.

For all training runs of \emph{1000} epochs, we set the initial learning rate to \emph{1e-3} and the batch size to \emph{3}. 
The networks are trained on two GPUs in parallel using \emph{6} random crops with a patch size of \emph{128x128x32} per batch element, with min/max normalization.

\clearpage
\noindent\textbf{\emph{DeepMACT (light-sheet microscopy)}}
For further validation, we evaluate the segmentation of micrometastasis in light-sheet microscopy images of the \emph{DeepMact} dataset \cite{pan2019deep}.
The volumes of \emph{350x350x350} resolution contain cleared tissue featuring different body parts of a mouse. 
We split the publicly available dataset into \emph{115} images for training and \emph{19} for testing.
The annotations feature a total of \emph{484} blobs, with \emph{6.99$\pm$8.14} blobs per sample.
As the data is sparsely annotated, \emph{F1} and especially \emph{DSC} should be interpreted with great care, as described in \Cref{sec:interpretation}.

For all training runs of \emph{500} epochs, we set the initial learning rate to \emph{1e-2} and the batch size to \emph{4}. 
The networks are trained on a single GPU using \emph{2} random crops with a patch size of \emph{192x192x48}.
The images are globally normalized based using a minimum and maximum threshold defined by the \emph{0.0} and \emph{99.5} percentile.

\clearpage
\section{Results}
\Cref{tab:main} summarizes the quantitative results of our experiments and \Cref{fig:volumetry,fig:detection} (Appendix) visualize qualitative results.
Across all datasets, we find that extending \emph{dice} to a \emph{blob loss} helps to improve detection metrics.
Furthermore, in some cases, we also observe improvements in volumetric performance measures.
While model selection seems not beneficial on this dataset, employing \emph{blob loss} produces more robust results, as both the \emph{dice} and \emph{tversky} models suffer performance drops for the \emph{best} checkpoints.
Notably, even though \emph{tversky} was explicitly proposed for MS lesion segmentation, it is clearly outperformed by \emph{dice}, as well as \emph{blob dice} and \emph{blob tversky}.
Further, even with the mitigation strategies suggested by the authors, \emph{inverse weighting} produced over-segmentations.

\begin{table}[ht]
\caption{
Experimental results for five datasets.
For all training runs with \emph{blob loss} we use $\alpha = 1$ and $\beta =2$.
Note that the results for LiTS are based on a different, more challenging test set and are therefore not comparable with the public leaderboard of the LiTS challenge.
For DISCO-MS, the \emph{dice} model completely over-segments and produces dissatisfactory results.
Therefore, we try two additional training runs with reduced learning rates following the heuristics suggested by \cite{bengio2012practical}, resulting in similar over-segmentation.
The same problem is observed for \emph{inverse weighting (iw)}.
\citet{shirokikh2020universal} themselves note the stability problems of the method and suggest lowering the learning rate to $1-e3$.
}
\nprounddigits{3}
\npdecimalsign{.}
\centering
\scriptsize
\begin{tblr}{colspec={lcr},colspec={|Q[2.5,c] Q[2.5,c] Q[1.0,c] | Q[1,c] Q[1,c] | Q[1,c] Q[1,c] Q[1,c] | Q[1,c] Q[1,c] Q[1,c]|}}
  \hline
  \textbf{dataset} & \textbf{loss} & \textbf{lr} & \textbf{DSC} & \textbf{SDSC} & \textbf{F1} & \textbf{IS} & \textbf{IP} \\
\hline
\SetCell[r=6]{m,1.5cm} MS
        &blob dice & 1e-2 & $\mathbf{\numprint{0.6800727681}}$ & $\mathbf{\numprint{0.8477524349}}$ & $\mathbf{\numprint{0.8099916811}}$ & $\numprint{0.8216573429}$ & $\mathbf{\numprint{0.827804315}}$ \\
        &dice & 1e-2 & $\numprint{0.6597225228}$ & $\numprint{0.8196077205}$ & $\numprint{0.7583294712}$ & $\mathbf{\numprint{0.8544719674}}$ & $\numprint{0.7106763497}$ \\
        &iw \cite{shirokikh2020universal} & 1e-2 & $\numprint{0.1528559143752418}$ & $\numprint{0.1668277320476198}$ & $\numprint{0.27819512044202993}$ & $\numprint{0.8012728861231966}$ & $\numprint{0.18818076597126945}$ \\
        &iw \cite{shirokikh2020universal} & 1e-3 & $\numprint{0.24330518125816972}$ & $\numprint{0.27275432856184256}$ & $\numprint{0.2822316891145946}$ & $\numprint{0.8187160381170797}$ & $\numprint{0.18865288439000966}$ \\
        \cline{2-8}
    
        &blob tversky & 1e-2 & $\mathbf{\numprint{0.6896689055}}$ & $\mathbf{\numprint{0.852262375}}$ & $\mathbf{\numprint{0.8037346995}}$ & $\numprint{0.8287599047}$ & $\mathbf{\numprint{0.8035478256}}$ \\
        &tversky & 1e-2 & $\numprint{0.6005395532}$ & $\numprint{0.6974265413}$ & $\numprint{0.565965955}$ & $\mathbf{\numprint{0.8538453315}}$ & $\numprint{0.4589202179}$ \\
\hline
\SetCell[r=2]{m,1.5cm} LiTS
        &blob dice & 1e-2 & $\mathbf{\numprint{0.6630027384}}$ & $\numprint{0.5423076852}$ & $\mathbf{\numprint{0.6570764578}}$ & $\mathbf{\numprint{0.8611963708}}$ & $\mathbf{\numprint{0.6310525232}}$ \\ 
        &dice & 1e-2 & $\numprint{0.6590172257}$ & $\mathbf{\numprint{0.5461734704}}$ & $\numprint{0.6225947326}$ & $\numprint{0.8011527978}$ & $\numprint{0.5991714217}$ \\ 
\hline
\SetCell[r=2]{m,1.5cm} SHANEL
        &blob dice & 1e-3 & $\mathbf{\numprint{0.5432052057}}$ & $\mathbf{\numprint{0.80782734}}$ & $\mathbf{\numprint{0.7919734904}}$ & $\mathbf{\numprint{0.873857404}}$ & $\mathbf{\numprint{0.7241205184}}$ \\
        &dice & 1e-3 & $\numprint{0.5394418734}$ & $\numprint{0.793608802}$ & $\numprint{0.7830215024}$ & $\numprint{0.8543570993}$ & $\numprint{0.7226804124}$ \\
\hline
\SetCell[r=4]{m,1.5cm} DISCO$-$MS
        &blob dice & 1e-3 & $\mathbf{\numprint{0.5456898724}}$ & $\mathbf{\numprint{0.6783278589}}$ & $\mathbf{\numprint{0.5894206549}}$ & $\numprint{0.7597402597}$ & $\mathbf{\numprint{0.4814814815}}$ \\ 
        
        &dice & 1e-3 & $\numprint{0.09523683643}$ & $\numprint{0.08298497289}$ & $\numprint{0.01239879713}$ & $\numprint{0.8701298701}$ & $\numprint{0.006243884255}$ \\

        &dice & 3e-4 & $\numprint{0.01585454894}$ & $\numprint{0.03616793724}$ & $\numprint{0.3791208791}$  & $\mathbf{\numprint{0.8961038961}}$ & $\numprint{0.2404181185}$ \\

        &dice & 1e-4 & $\numprint{0.00747126907}$ & $\numprint{0.01118405106}$ & $\numprint{0.2278026906}$  & $\numprint{0.8246753247}$ & $\numprint{0.1321540062}$ \\
\hline
\SetCell[r=2]{m,1.5cm} DeepMACT        
        &blob dice & 1e-2 & $\mathbf{\numprint{0.3572976835}}$ & $\mathbf{\numprint{0.3927128197}}$ & $\mathbf{\numprint{0.3909419599}}$ & $\mathbf{\numprint{0.8711931344}}$ & $\mathbf{\numprint{0.275945358}}$ \\ 
        &dice & 1e-2 & $\numprint{0.3525346754}$ & $\numprint{0.3715205018}$ & $\numprint{0.3665096197}$ & $\numprint{0.8014164198}$ & $\numprint{0.2540287764}$ \\
\hline
\end{tblr}
\label{tab:main}
\end{table}

\begin{table*}[ht!]
\caption{
Ablation analysis on the \emph{blob loss'} hyperparameters $\alpha$ and $\beta$ for the MS lesions dataset.
We observe that \emph{blob loss} seems to be quite robust with regard to hyperparameter choice, as long as the global term remains present, compare \Cref{eq:whole}.
The default parameters $\alpha = 2$ and $\beta = 1$ provide the best results.
}
\nprounddigits{3}
\npdecimalsign{.}
\centering
\scriptsize
\begin{tblr}{colspec={lcr},colspec={|Q[2.0,c] Q[0.5,c] Q[0.5,c] | Q[1,c] Q[1,c] Q[1,c] Q[1,c] | Q[1,c] Q[1,c] Q[1,c]|}}
  \hline
\textbf{loss} & $\mathbf{\alpha}$ & $\mathbf{\beta}$ & \textbf{DSC} & \textbf{S} & \textbf{P} & \textbf{SDSC} & \textbf{F1} & \textbf{IS} & \textbf{IP} \\
\hline
        blob dice & 3 & 1 & $\numprint{0.6735914934}$ & $\numprint{0.6290213222399}$ & $\numprint{0.764687116849168}$ & $\numprint{0.832755501875923}$ & $\numprint{0.790058709882785}$ & $\numprint{0.796372540447021}$ & $\numprint{0.815062607977183}$ \\ 
        blob dice & 2 & 1 & $\mathbf{\numprint{0.6800727681}}$ & $\numprint{0.6256513266}$ & $\numprint{0.7821899105}$ & $\mathbf{\numprint{0.8477524349}}$ & $\mathbf{\numprint{0.8099916811}}$ & $\numprint{0.8216573429}$ & $\mathbf{\numprint{0.827804315}}$ \\ 
        blob dice &1 & 1 & $\numprint{0.6576324384}$ & $\numprint{0.5804289091}$ & $\numprint{0.8018632439}$ & $\numprint{0.8388776647}$ & $\numprint{0.804209065}$ & $\numprint{0.8402097852}$ & $\numprint{0.8005730304}$ \\ 
        blob dice &1 & 2 & $\numprint{0.6300293434}$ & $\numprint{0.5518268732}$ & $\numprint{0.8025617145}$ & $\numprint{0.8187564386}$ & $\numprint{0.792328637}$ & $\numprint{0.8323627337}$ & $\numprint{0.7861103432}$ \\ 
        dice &1 & 0 & $\numprint{0.6597225228}$ & $\mathbf{\numprint{0.7036249179}}$ & $\numprint{0.6563665925}$ & $\numprint{0.8196077205}$ & $\numprint{0.7583294712}$ & $\mathbf{\numprint{0.8544719674}}$ & $\numprint{0.7106763497}$ \\ 
        blob & 0 & 1 &  $\numprint{0.5216279391}$ & $\numprint{0.4088525507}$ & $\mathbf{\numprint{0.8366344888}}$ & $\numprint{0.7276010605}$ & $\numprint{0.7437368737}$ & $\numprint{0.8054148135}$ & $\numprint{0.7271671112}$ \\
   \hline
\end{tblr}
\label{tab:ablation}
\end{table*}

\Cref{tab:ablation} summarizes the results of the ablation study on $\alpha$ and $\beta$ parameters of \emph{blob loss}.
We find that assigning higher importance to the global parameter by choosing $\alpha$ = \emph{2} and $\beta$ = \emph{1} seems to produce the best results.
Overall, we find that \emph{blob loss} seems quite robust regarding the choice of hyperparameters as long as the global term remains included by choosing a $\alpha$ greater than \emph{0}.

\clearpage
\section{Discussion}
\noindent\textbf{Contribution:} 
\emph{blob loss} can be employed to provide existing loss functions with instance imbalance awareness.
We demonstrate that the application of \emph{blob loss} improves detection- and in some cases, even volumetric segmentation performance across a diverse set of complex 3D bio-medical segmentation tasks.
We evaluate \emph{blob loss}' performance in the segmentation of multiple sclerosis (MS) lesions in MR, liver tumors in CT, and segmentation of different biological structures in 3D light-sheet microscopy datasets.
Depending on the dataset, it achieves these improvements either due to better detection of foreground objects, better suppression of background objects, or both.
We provide an implementation of blob loss leveraging on a precomputed connected component analysis for fast processing times.

\noindent\textbf{Limitations:}
Certainly, the biggest disadvantage of \emph{blob loss} is the dependency on instance segmentation labels; however, in many cases, these can be simply obtained by a connected component analysis, as demonstrated in our experiments.
Another disadvantage of \emph{blob loss} compared to other loss functions are the more extensive computational requirements.
By definition, the user is required to run computations with large patch sizes that feature multiple instances.
This results in an increased demand for GPU memory, especially when working with 3D data (as in our experiments).
However, larger patch sizes have proven helpful for bio-medical segmentation problems, in general, \cite{isensee2021nnu}.
Furthermore, according to our formulation, \emph{blob loss} possesses an interesting mathematical property,
it penalizes false positives proportionally to the number of instances in the volume.
Additionally, even though \emph{blob loss} can easily be reduced to a single hyperparameter, and it proved quite robust in our experiments, it might be sensitive to hyperparameter tuning.
Moreover, by design \emph{blob loss} can only improve performance for multi-instance segmentation problems.

\noindent\textbf{Interpretation:}
One can only speculate why \emph{blob loss} improves performance metrics.
CNNs learn features that are very sensitive to texture \cite{geirhos2018imagenet}.
Unlike conventional loss functions, \emph{blob loss} adds attention to every single instance in the volume.
Thus the network is forced to learn the instance imbalanced features such as, but not limited to morphology and texture, which would not be well represented by optimizing via \emph{dice} and alike.
Such instance imbalance was observed in the medical field, as it has been shown that MS lesions change their imaging phenotype over time, with recent lesions looking significantly different from older ones \cite{elliott2019slowly}.
These aspects might explain the gains in instance sensitivity.
Furthermore, adding the multiple instance terms leads to heavy penalization on background, which might explain why we often observe an improvement in precision, see supplementary materials.

\noindent\textbf{Outlook:}
Future research will have to reveal to which extent transformation to \emph{blob loss} can be beneficial for other segmentation tasks and loss functions.
A first and third place in recent public segmentation challenges using a compound-based variant \emph{blob loss} indicate that \emph{blob loss} might possess broad applicability towards other instance imbalanced semantic segmentation problems.

\vspace{1.0cm}

\clearpage
\FloatBarrier
\section*{Acknowledgement}
\noindent Bjoern Menze, Benedikt Wiestler and Florian Kofler are supported through the SFB 824, subproject B12.

\noindent Supported by Deutsche Forschungsgemeinschaft (DFG) through TUM International Graduate School of Science and Engineering (IGSSE), GSC 81.

\noindent Lucas Fidon, Suprosanna Shit and Ivan Ezhov are supported by the Translational Brain Imaging Training Network(TRABIT) under the European Union's `Horizon 2020' research \& innovation program (Grant agreement ID: 765148).

\noindent Ivan Ezhov, Suprosanna Shit are funded by DComEX (Grant agreement ID: 956201).

\noindent With the support of the Technical University of Munich – Institute for Advanced Study, funded by the German Excellence Initiative.

\noindent Supported by Anna Valentina Lioba Eleonora Claire Javid Mamasani.

\noindent Suprosanna Shit is supported by the Graduate School of Bioengineering,  Technical University of Munich.

\noindent Jan Kirschke has received Grants from the ERC, DFG, BMBF and is Co-Founder of Bonescreen GmbH.

\noindent Bjoern Menze acknowledges support by the Helmut Horten Foundation.

\noindent Research reported in this publication was partly supported by AIME GPU cloud services.

\clearpage
\setcounter{figure}{41}
\setcounter{table}{41}

\section{Supplementary Materials}
In the supplementary materials we use the acronyms introduced in the main text.
For easier understanding we repeat the ones used in the tables:
learning-rate (lr); Sørensen-Dice similarity coefficient (DSC); Surface Dice similarity coefficient (SDSC); Instance sensitivity (IS); Instance Precision (IP)

\subsection{Data Set Description}

%
\begin{figure}[h!]
    \centering
    \includegraphics[width=0.55\linewidth]{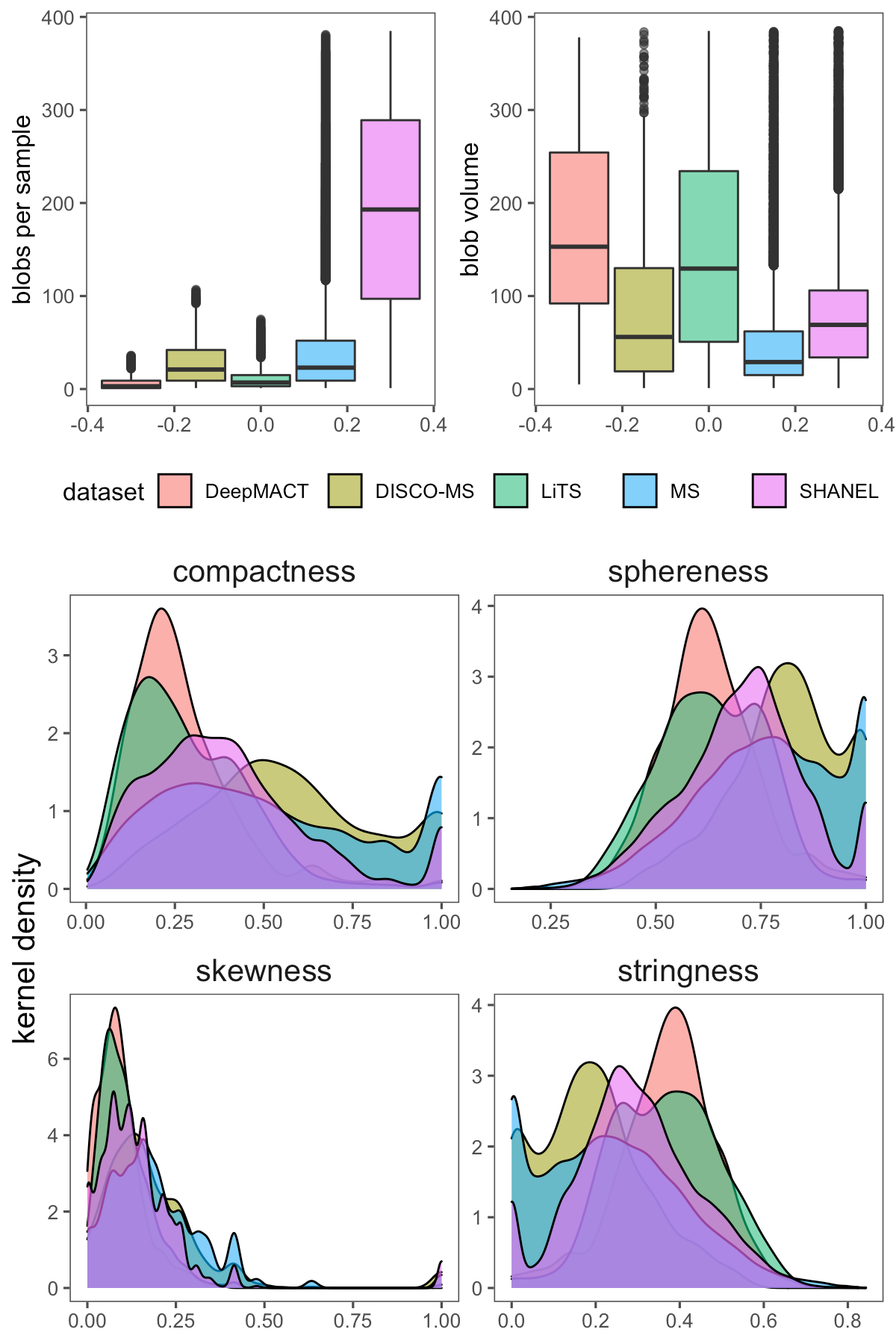}
    \caption{
    Dataset heterogeneity and instance imbalance.
    Boxplot left: blob count per sample;
    Boxplot right: volume per blob.
    For visualization purposes, the boxplots are cut off at the \emph{95\%} percentile.
    The four kernel density plots on the bottom depict four shape features, defined as:
    \emph{Compactness}: Volume of blob divided by the volume of a hypothetical enclosing sphere.
    \emph{Sphereness}: Ratio of the maximum distance within a blob to the diameter of a hypothetical sphere with the same volume as the blob.
    \emph{Stringness}: Equals $1-Sphereness$ and approaches 1 for string-like shapes.
    \emph{Skewness}:  Approaches \emph{1} if a blob is thick or dense on one end and has a large tail on the other side.    
    }
    \label{fig:datasets}
\end{figure}
\FloatBarrier

\subsection{Data Availability Statements}
\noindent As the MS dataset contains sensitive patient information, requests for the MS data will be reviewed by an ethical committee to determine whether the request is subject to
any confidentiality obligations.
Requests should contain a research proposal, an ethics statement, and a data transfer agreement.

\noindent LiTS challenge data is publicly available at
\href{https://competitions.codalab.org/competitions/17094#participate}{codalab.org}

\noindent SHANEL, DISCO-MS, and DeepMACT datasets are available upon request via  \href{http://discotechnologies.org/}{discotechnologies.org}

\FloatBarrier


\subsection{Why the two hyper-parameters $\alpha$ and $\beta$?}
One might wonder why \emph{blob loss} is defined with $\alpha$ and $\beta$, while $\beta$ could easily be defined as $1-\alpha$.
Assume one would want to extend a conventional soft dice loss (this equals $\alpha=1$ and $\beta=0$) to a \emph{blob dice} with equal weights to see if it aids segmentation performance.
In this case, $\alpha$ and $\beta$ would both become $0.5$.
As the initial soft dice loss is now halved, one needs to double the learning rate to maintain comparability between training runs.
These learning rate adjustments can become cumbersome, especially when operating with non equal weighted loss candidates.
\FloatBarrier

\subsection{Why do we need masking?}
To explore the effect of the masking defined in \Cref{eq:masking} we train another model on the MS dataset.
Therefore, we keep all training parameters constant except for the loss.
Here, we skip the masking inside the loss computation.
We find that the masking is crucial for achieving a high segmentation performance, see \Cref{tab:mask}.

\begin{table}[htbp]
\caption{
   Segmentation performance when computing the loss without masking.
   The network tends to oversegment as reflected by the high instance sensitivity (IS) compared to the low instance precision (IP).
   The low performance is also reflected by the volumetric metrics DSC and SDSC.
    }
\centering
\nprounddigits{3}
\npdecimalsign{.}
\begin{tblr}{colspec={lcr},colspec={|Q[1.5,c] Q[2.5,c] Q[2.5,c] Q[1,c] | Q[1,c] Q[1,c] Q[1,c] Q[1,c] Q[1,c]|}}
  \hline
\textbf{dataset} & \textbf{architecture} & \textbf{loss} & \textbf{lr} & \textbf{DSC} & \textbf{SDSC} & \textbf{F1} & \textbf{IS} & \textbf{IP} \\ 
  \hline
MS & \href{https://docs.monai.io/en/stable/networks.html\#basicunet}{\emph{BasicUnet}} &blob dice & 1e-2 & $\mathbf{\numprint{0.6800727681}}$ & $\mathbf{\numprint{0.8477524349}}$ & $\mathbf{\numprint{0.8099916811}}$ & $\numprint{0.8216573429}$ & $\mathbf{\numprint{0.827804315}}$ \\
MS & \href{https://docs.monai.io/en/stable/networks.html\#basicunet}{\emph{BasicUnet}} &dice & 1e-2 & $\numprint{0.6597225228}$ & $\numprint{0.8196077205}$ & $\numprint{0.7583294712}$ & $\mathbf{\numprint{0.8544719674}}$ & $\numprint{0.7106763497}$ \\
MS & \href{https://docs.monai.io/en/stable/networks.html\#basicunet}{\emph{BasicUnet}} & no\_masking &   1e-2 &   0.358 & 0.382 & 0.297 &  0.829 & 0.201 \\  
   \hline
\end{tblr}
\label{tab:mask}
\end{table}

\FloatBarrier

\vspace{2cm}
\subsection{Qualitative Segmentation Performance}

%
\begin{figure}[h!]
    \centering
    \includegraphics[width=1.0\linewidth]{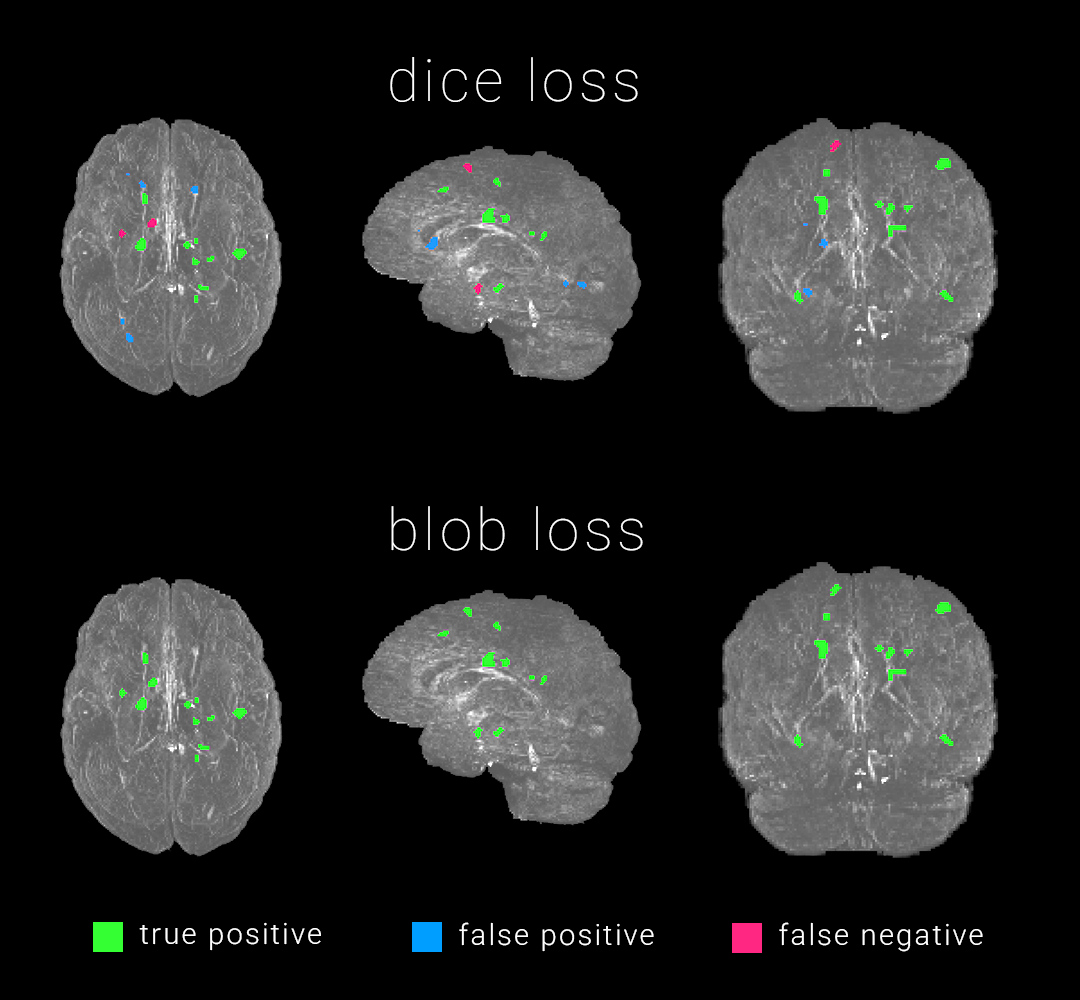}
    \caption{
    Comparison of detection performance.
    Maximum intensity projections of the FLAIR images overlayed with segmentations for \emph{dice} and \emph{blob dice}.
    Lesions are colored according to their detection status:
    Green for \emph{true positive};
    Blue for \emph{false positive};
    Red for \emph{false negative}.
    In this case, transformation to a \emph{blob loss} improves \emph{F1} from \emph{0.74} to \emph{1.0}.
    This is driven by an increase in \emph{instance-sensitivity} from \emph{0.83} to \emph{1.0} and \emph{instance-precision} from \emph{0.67} to \emph{1.0}.
    Therefore, for this particular patient \emph{blob loss} boosts \emph{F1} by simultaneously improving detection of foreground and suppression of background signal.
    This seems to be a regular pattern across multiple segmentation problems, compare \Cref{tab:main}.
    Refer to \Cref{fig:volumetry} for a volumetric analysis.
    }
    \label{fig:detection}
\end{figure}
%
%
\begin{figure}[ht!]
    \centering
    \includegraphics[width=0.95\linewidth]{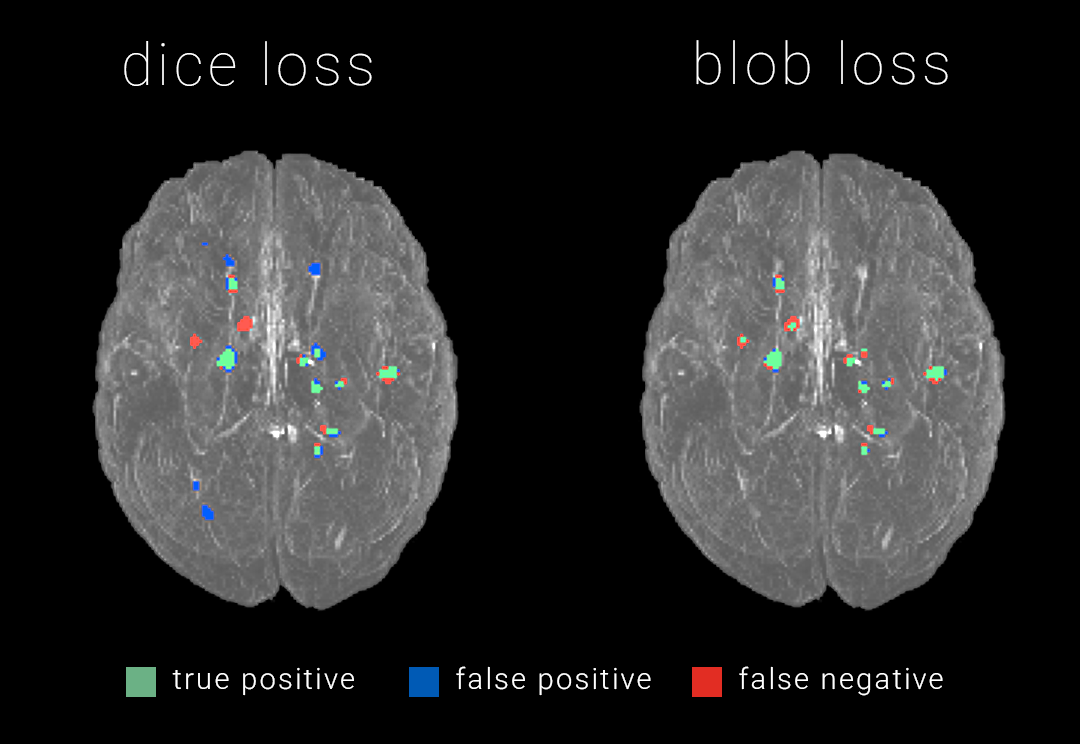}
    \caption{
    Comparison of volumetric segmentation performance.
    Maximum intensity projections of the FLAIR images overlayed with segmentations for \emph{dice} and \emph{blob dice}.
    Lesions are colored according to their detection status:
    Green for \emph{true positive};
    Blue for \emph{false positive};
    Red for \emph{false negative}.
    For this particular patient, applying the transformation to a \emph{blob loss} improves the volumetric Dice coefficient from \emph{0.56} to \emph{0.70}.
    This is caused by an increase in \emph{volumetric precision} from \emph{0.48} to \emph{0.75}, while the 
    \emph{volumetric sensitivity} remains constant at \emph{0.66}.
    Thus, unlike on the detection level, the improvement is solely reached due to better suppression of background signal,  compare \Cref{fig:detection}.
    }
    \label{fig:volumetry}
\end{figure}
%
\FloatBarrier

\subsection{MS model selection experiment}

\begin{table}[ht!]
\caption{
Results for the model selection experiment on MS data.
We evaluate the checkpoint with the lowest loss on the validation set of \emph{21} patients.
As \emph{dice} and especially \emph{tversky} suffer big performance drops, conducting model selection seems to hurt the networks' generalization ability.
Training with \emph{blob loss} leads to more robust performance. 
}
\nprounddigits{3}
\npdecimalsign{.}
\centering
\begin{tblr}{colspec={lcr},colspec={|Q[1.5,c] Q[1.5,c] Q[2.0,c] | Q[0.75,c] Q[0.75,c] Q[0.75,c] Q[0.75,c] Q[0.75,c]|}}
  \hline
\textbf{dataset} & \textbf{architecture} & \textbf{loss} & \textbf{DSC} & \textbf{SDSC} & \textbf{F1} & \textbf{IS} & \textbf{IP} \\ 
  \hline
MS & \href{https://docs.monai.io/en/stable/networks.html\#basicunet}{\emph{BasicUnet}} & blob dice & $\mathbf{\numprint{0.6776580764}}$ & $\mathbf{\numprint{0.8490951296}}$ & $\mathbf{\numprint{0.80028352}}$ & $\mathbf{\numprint{0.8594021584}}$ & $\mathbf{\numprint{0.7745127965}}$ \\
MS & \href{https://docs.monai.io/en/stable/networks.html\#basicunet}{\emph{BasicUnet}} & dice & $\numprint{0.640074436}$ & $\numprint{0.800085831}$ & $\numprint{0.7592185379}$ & $\numprint{0.8101243149}$ & $\numprint{0.7479924018}$ \\
\hline
MS & \href{https://docs.monai.io/en/stable/networks.html\#basicunet}{\emph{BasicUnet}} & blob tversky & $\mathbf{\numprint{0.6917393939}}$ & $\mathbf{\numprint{0.8571838183}}$ & $\mathbf{\numprint{0.7960696753}}$ & $\numprint{0.8484746025}$ & $\mathbf{\numprint{0.7752554198}}$ \\
MS & \href{https://docs.monai.io/en/stable/networks.html\#basicunet}{\emph{BasicUnet}} & tversky & $\numprint{0.4051501989}$ & $\numprint{0.4608221263}$ & $\numprint{0.4127952906}$ & $\mathbf{\numprint{0.8508559125}}$ & $\numprint{0.3024063971}$ \\
   \hline
\end{tblr}
\label{tab:select}
\end{table}

\FloatBarrier

\FloatBarrier
\clearpage
\subsection{Multi-class Segmentation Extension of \textbf{\emph{blob loss}}}
Extension to multi-class segmentation problems can be achieved by summing across the foreground classes.
Let $\mathcal{L}$ a loss function for binary segmentation, and let $\mathbf{p}=(p_i^c)_{i\in \Omega, c=0\ldots C}$ be the predicted segmentation and $\mathbf{g}=(g_i^c)_{i\in \Omega, c=0\ldots C}$ the one-hot encoding of the ground-truth segmentation, where $\Omega$ is the image domain and $C\geq 1$ is the number of foreground classes.
We assume without loss of generality that the background corresponds to the class $c=0$.

We propose to define the instance-aware loss function $\mathcal{L}_{blob}$ associated with $\mathcal{L}$ as
\begin{equation}
    \mathcal{L}_{blob}\left(\mathbf{p}, \mathbf{g}\right)
    = 
    \frac{1}{C}
    \sum_{c=1}^C
    \left(
    \frac{1}{N_c}
    \sum_{n=1}^{N_c} \mathcal{L}
    \left(
    (p_i^c)_{i\in \Omega_{c,n}}, (g_i^c)_{i\in \Omega_{c,n}}
    \right)
    \right)
    \label{eq:mc_masking}
\end{equation}
where $\Omega_{c,n}$ is the image domain after excluding the voxels labeled as $c$ in the ground-truth segmentation, and that does not belong to the instance $n$.

The final loss for multi-class segmentation has the same form as for the binary segmentation case
\begin{equation}
\mathcal{L}  = \alpha \mathcal{L}_{global} + \beta \mathcal{L}_{blob}
\label{eq:mc_whole}
\end{equation}
where $\alpha \geq 0$ and $\beta \geq 0$.
Depending on the segmentation task at hand, one could further consider the introduction of class-specific $\alpha$ and $\beta$.

\FloatBarrier

\subsection{Dependency on Network architecture}
\begin{figure*}[h]
    \centering
    \includegraphics[width=0.98\textwidth]{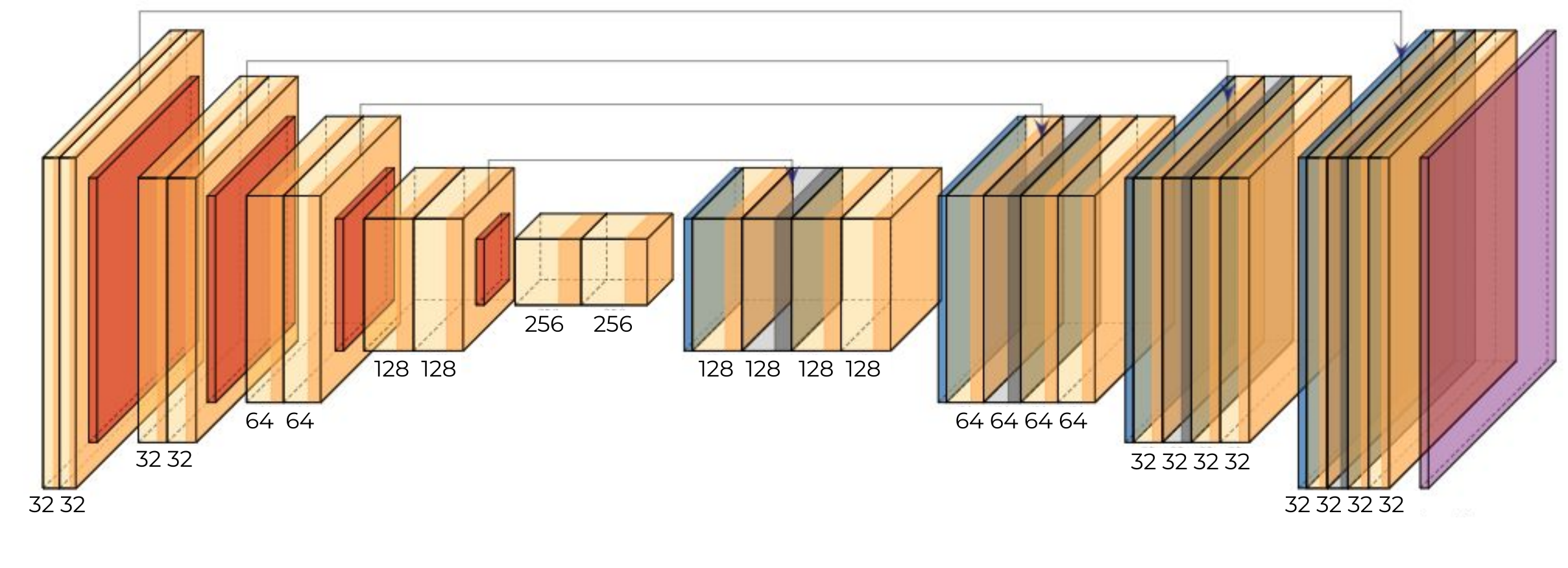}
    \caption{The U-Net architecture used in the experiments, consisting of encoder and decoder parts.
    The numbers under the convolutional layers represent the channel count.
    The implementation is available via \href{https://docs.monai.io/en/0.7.0/_modules/monai/networks/nets/basic_unet.html}{MONAI} and is inspired by \citet{falk2019u}.
    }
    \label{fig:unet}
\end{figure*}

U-Nets are the state-of-the-art architecture for bio-medical semantic segmentation \cite{ronneberger2015u}.
As our primary goal is to evaluate a novel loss function, we choose the most standard U-Net architecture we could find for our experiments, see \Cref{fig:unet}.
To investigate whether blob loss is dependent on the architecture we conduct a network architecture ablation study on the MS dataset.
Despite exchanging the network architecture, we set the learning rate to $1e-3$ and keep the other parameters constant to the previous training runs.
We compare a \emph{blob dice} and \emph{dice} dice baseline with \emph{UNETR} \cite{hatamizadeh2022unetr}.
Unlike our basic U-Net implementation \emph{UNETR} features transformers.
The implementation is available via \href{https://docs.monai.io/en/0.7.0/networks.html\#unetr}{MONAI}.
\Cref{tab:architecture} summarizes the results of the model selection ablation study.

\begin{table}[htbp]
\caption{
Architecture ablation study.
We compare \emph{dice} against \emph{blob dice} for training runs with \emph{UNETR} \cite{hatamizadeh2022unetr}.
The results indicate that the performance improvements of \emph{blob loss} are architecture agnostic.
}
\centering
\begin{tblr}{colspec={lcr},colspec={|Q[1.5,c] Q[2.5,c] Q[2.5,c] Q[1,c] | Q[1,c] Q[1,c] Q[1,c] Q[1,c] Q[1,c]|}}
  \hline
\textbf{dataset} & \textbf{architecture} & \textbf{loss} & \textbf{lr} & \textbf{DSC} & \textbf{SDSC} & \textbf{F1} & \textbf{IS} & \textbf{IP} \\ 
  \hline
MS & \href{https://arxiv.org/abs/2103.10504}{\emph{UNETR}} & dice &   1e-3 &   0.380 &   0.386 & 0.383 &  0.870 & 0.272 \\ 
MS & \href{https://arxiv.org/abs/2103.10504}{\emph{UNETR}} & blob dice &   1e-3 &   \textbf{0.632} & \textbf{0.789} &   \textbf{0.691} &  \textbf{0.889} & \textbf{0.600} \\ 
   \hline
\end{tblr}
\label{tab:architecture}
\end{table}

\FloatBarrier

\subsection{Increased Instance-wise Penalization of False Positives for Soft Dice Loss}
Applying \emph{blob loss} leads to increased instance-wise penalization of false positives.
Let $Y$ be the predicted segmentation, $L$ be the ground-truth binary segmentation, and $N>0$ the number of instances. Reasonably, we can assume that our ground truth is perfect and can be used for accurate false-positives and false-negatives categorization.
Let $\{L_i\}_{i=1}^{N}$ be the masks of the binary instances. Hence, by construction we have $\sum_{i=1}^NL_i=L$. Further, we define the domain mask for each instance $L_i$ as follows
\begin{equation}
    \Omega_i = I - \sum_{j=1,\,j\neq i}^N L_j
\end{equation}
where $I$ is the mask on the image domain $\Omega$ with all its values equal to $1$.

Although we predict all instances in a single channel, one can decompose it into $N+1$ number of channel
\begin{equation}
    Y = \sum_{i=1}^NL_i\odot Y + \left(I-\sum_{i=1}^NL_i\right)\odot Y
\end{equation}
where $\odot$ denotes Hadamard product. We denote $L_i\odot Y$ as $Y_i$ and $(I-\sum_{i=1}^NL_i)\odot Y$ as $Y_{fp}$. Note that $Y_i$ consists of voxel-wise true-positives and $Y_{fp}$ consists of false-positives. Using the definition of the blob loss in \Cref{eq:masking}, with $\mathcal{L}$ the soft Dice loss, $g= L_i$ and $p= Y\odot\Omega_i$
\begin{align}
    \mathcal{L}
    \left(
    (p^j)_{j\in \Omega_i}, (g^j)_{j\in \Omega_i}
    \right) 
    &=\dfrac{2\times \sum_{j\in \Omega_i}g^j p^j}{\sum_{j\in \Omega_i}g_i^j+\sum_{j\in \Omega_i}p_i^j} \nonumber ;~[\mbox{superscript $j$ denotes each voxel}]\\
    & = \dfrac{2\times \sum_{j=1}^{M}L_i^j Y^j \Omega_i^j}{\sum_{j=1}^{M}L_i^j+\sum_{j=1}^{M}(Y^j\odot\Omega_i^j)}\nonumber\\
    & = \dfrac{2\times \sum_{j=1}^{M}(L_i \odot Y)^j}{\sum_{j=1}^{M}L_i^j+\sum_{j=1}^{M}(Y\odot\Omega_i)^j}; ~[\mbox{using the fact $L_i\odot \Omega_i=L_i$}\label{eq:diceomega}]
\end{align}
considering only $Y\odot\Omega_i$ we have
\begin{align}
    Y\odot\Omega_i &= Y\odot\Omega_i\odot L_i +  Y\odot\Omega_i\odot(I-L_i))\nonumber\\
     & = Y\odot L_i+ Y\odot(I-\sum_{i=1,i\neq j}^NL_j)\odot(I-L_i)\nonumber\\
     & = Y\odot L_i+Y\odot(I-\sum_{i=1,i\neq j}^NL_j) -Y\odot L_i\nonumber; ~ [\mbox{since $L_i \odot L_j=0$ for $i\neq j$}]\\
     & = Y\odot L_i+Y\odot(I-\sum_{i=1}^NL_j)\nonumber\\
     & = \sum_{j=1}^{M}Y_i^j+\sum_{j=1}^{M}Y^j_{fp}\label{eq:ygamma}
\end{align}
Replacing Eq. \ref{eq:ygamma} in Eq. \ref{eq:diceomega} we get
\begin{align}
     \mathcal{L}
    \left(
    (p^j)_{j\in \Omega_i}, (g^j)_{j\in \Omega_i}
    \right)
    &= \dfrac{2\times \sum_{j=1}^{M}Y_i^j}{\sum_{j=1}^{M}L_i^j+\sum_{j=1}^{M}Y_i^j+\sum_{j=1}^{M}Y^j_{fp}}
\end{align}
The term $\sum_{j=1}^{M}Y^j_{fp}$ in the denominator penalizes the false positive and is present for every instance $i$.
Therefore, the instance-aware soft Dice loss penalizes more the false positives than the soft Dice loss. Also note that the degree of false-positives penalty is proportional to the number of instances. This dynamic penalty is beneficial in preventing network from over-predicting in case of many instances while it does not affect much for fewer instances.

\clearpage




%
%
%

\bibliography{flow/long_references.bib}

\begin{thebibliography}{33}
\providecommand{\natexlab}[1]{#1}
\providecommand{\url}[1]{\texttt{#1}}
\providecommand{\urlprefix}{URL }
\expandafter\ifx\csname urlstyle\endcsname\relax
  \providecommand{\doi}[1]{doi:\discretionary{}{}{}#1}\else
  \providecommand{\doi}{doi:\discretionary{}{}{}\begingroup
  \urlstyle{rm}\Url}\fi

\bibitem[{Bengio(2012)}]{bengio2012practical}
Bengio, Y.: Practical recommendations for gradient-based training of deep
  architectures (2012)

\bibitem[{Berman et~al.(2018)Berman, Triki, and Blaschko}]{berman2018lovasz}
Berman, M., Triki, A.R., Blaschko, M.B.: The lov{\'a}sz-softmax loss: A
  tractable surrogate for the optimization of the intersection-over-union
  measure in neural networks. In: Proceedings of the IEEE Conference on
  Computer Vision and Pattern Recognition, pp. 4413--4421 (2018)

\bibitem[{Bhatia et~al.(2021)Bhatia, Brunner, Rong, Mai, Thielert, Al-Maskari,
  Paetzold, Kofler, Todorov, Ali et~al.}]{bhatia2021proteomics}
Bhatia, H., Brunner, A., Rong, Z., Mai, H., Thielert, M., Al-Maskari, R.,
  Paetzold, J., Kofler, F., Todorov, M., Ali, M., et~al.: Proteomics of
  spatially identified tissues in whole organs. arXiv  (2021)

\bibitem[{Bilic et~al.(2019)Bilic, Christ, Vorontsov, Chlebus, Chen, Dou, Fu,
  Han, Heng, Hesser, Kadoury, Konopczynski, Le, Li, Li, Lipkovà, Lowengrub,
  Meine, Moltz, Pal, Piraud, Qi, Qi, Rempfler, Roth, Schenk, Sekuboyina,
  Vorontsov, Zhou, Hülsemeyer, Beetz, Ettlinger, Gruen, Kaissis, Lohöfer,
  Braren, Holch, Hofmann, Sommer, Heinemann, Jacobs, Mamani, van Ginneken,
  Chartrand, Tang, Drozdzal, Ben-Cohen, Klang, Amitai, Konen, Greenspan,
  Moreau, Hostettler, Soler, Vivanti, Szeskin, Lev-Cohain, Sosna, Joskowicz,
  and Menze}]{bilic2019liver}
Bilic, P., Christ, P.F., Vorontsov, E., Chlebus, G., Chen, H., Dou, Q., Fu,
  C.W., Han, X., Heng, P.A., Hesser, J., Kadoury, S., Konopczynski, T., Le, M.,
  Li, C., Li, X., Lipkovà, J., Lowengrub, J., Meine, H., Moltz, J.H., Pal, C.,
  Piraud, M., Qi, X., Qi, J., Rempfler, M., Roth, K., Schenk, A., Sekuboyina,
  A., Vorontsov, E., Zhou, P., Hülsemeyer, C., Beetz, M., Ettlinger, F.,
  Gruen, F., Kaissis, G., Lohöfer, F., Braren, R., Holch, J., Hofmann, F.,
  Sommer, W., Heinemann, V., Jacobs, C., Mamani, G.E.H., van Ginneken, B.,
  Chartrand, G., Tang, A., Drozdzal, M., Ben-Cohen, A., Klang, E., Amitai,
  M.M., Konen, E., Greenspan, H., Moreau, J., Hostettler, A., Soler, L.,
  Vivanti, R., Szeskin, A., Lev-Cohain, N., Sosna, J., Joskowicz, L., Menze,
  B.H.: The liver tumor segmentation benchmark (lits) (2019)

\bibitem[{Caicedo et~al.(2019)Caicedo, Goodman, Karhohs, Cimini, Ackerman,
  Haghighi, Heng, Becker, Doan, McQuin et~al.}]{caicedo2019nucleus}
Caicedo, J.C., Goodman, A., Karhohs, K.W., Cimini, B.A., Ackerman, J.,
  Haghighi, M., Heng, C., Becker, T., Doan, M., McQuin, C., et~al.: Nucleus
  segmentation across imaging experiments: the 2018 data science bowl. Nature
  methods \textbf{16}(12), 1247--1253 (2019)

\bibitem[{Eelbode et~al.(2020)Eelbode, Bertels, Berman, Vandermeulen, Maes,
  Bisschops, and Blaschko}]{eelbode2020optimization}
Eelbode, T., Bertels, J., Berman, M., Vandermeulen, D., Maes, F., Bisschops,
  R., Blaschko, M.B.: Optimization for medical image segmentation: theory and
  practice when evaluating with dice score or jaccard index. IEEE Transactions
  on Medical Imaging \textbf{39}(11), 3679--3690 (2020)

\bibitem[{Elliott et~al.(2019)Elliott, Wolinsky, Hauser, Kappos, Barkhof,
  Bernasconi, Wei, Belachew, and Arnold}]{elliott2019slowly}
Elliott, C., Wolinsky, J.S., Hauser, S.L., Kappos, L., Barkhof, F., Bernasconi,
  C., Wei, W., Belachew, S., Arnold, D.L.: Slowly expanding/evolving lesions as
  a magnetic resonance imaging marker of chronic active multiple sclerosis
  lesions. Multiple Sclerosis Journal \textbf{25}(14), 1915--1925 (2019)

\bibitem[{Falk et~al.(2019)Falk, Mai, Bensch, {\c{C}}i{\c{c}}ek, Abdulkadir,
  Marrakchi, B{\"o}hm, Deubner, J{\"a}ckel, Seiwald et~al.}]{falk2019u}
Falk, T., Mai, D., Bensch, R., {\c{C}}i{\c{c}}ek, {\"O}., Abdulkadir, A.,
  Marrakchi, Y., B{\"o}hm, A., Deubner, J., J{\"a}ckel, Z., Seiwald, K.,
  et~al.: U-net: deep learning for cell counting, detection, and morphometry.
  Nature methods \textbf{16}(1), 67--70 (2019)

\bibitem[{Fidon et~al.(2017)Fidon, Li, Garcia-Peraza-Herrera, Ekanayake,
  Kitchen, Ourselin, and Vercauteren}]{fidon2017generalised}
Fidon, L., Li, W., Garcia-Peraza-Herrera, L.C., Ekanayake, J., Kitchen, N.,
  Ourselin, S., Vercauteren, T.: Generalised wasserstein dice score for
  imbalanced multi-class segmentation using holistic convolutional networks.
  In: International MICCAI Brainlesion Workshop, pp. 64--76, Springer (2017)

\bibitem[{Geirhos et~al.(2018)Geirhos, Rubisch, Michaelis, Bethge, Wichmann,
  and Brendel}]{geirhos2018imagenet}
Geirhos, R., Rubisch, P., Michaelis, C., Bethge, M., Wichmann, F.A., Brendel,
  W.: Imagenet-trained cnns are biased towards texture; increasing shape bias
  improves accuracy and robustness. arXiv preprint arXiv:1811.12231  (2018)

\bibitem[{Hatamizadeh et~al.(2022)Hatamizadeh, Tang, Nath, Yang, Myronenko,
  Landman, Roth, and Xu}]{hatamizadeh2022unetr}
Hatamizadeh, A., Tang, Y., Nath, V., Yang, D., Myronenko, A., Landman, B.,
  Roth, H.R., Xu, D.: Unetr: Transformers for 3d medical image segmentation.
  In: Proceedings of the IEEE/CVF Winter Conference on Applications of Computer
  Vision, pp. 574--584 (2022)

\bibitem[{He et~al.(2017)He, Gkioxari, Doll{\'a}r, and Girshick}]{he2017mask}
He, K., Gkioxari, G., Doll{\'a}r, P., Girshick, R.: Mask r-cnn. In: Proceedings
  of the IEEE international conference on computer vision, pp. 2961--2969
  (2017)

\bibitem[{Isensee et~al.(2021)Isensee, Jaeger, Kohl, Petersen, and
  Maier-Hein}]{isensee2021nnu}
Isensee, F., Jaeger, P.F., Kohl, S.A., Petersen, J., Maier-Hein, K.H.: nnu-net:
  a self-configuring method for deep learning-based biomedical image
  segmentation. Nature methods \textbf{18}(2), 203--211 (2021)

\bibitem[{Isensee et~al.(2019)Isensee, Petersen, Kohl, J{\"a}ger, and
  Maier-Hein}]{isensee2019nnu}
Isensee, F., Petersen, J., Kohl, S.A., J{\"a}ger, P.F., Maier-Hein, K.H.:
  nnu-net: Breaking the spell on successful medical image segmentation. arXiv
  preprint arXiv:1904.08128 \textbf{1}, 1--8 (2019)

\bibitem[{Jungo and et~al(2021)}]{jungo2021pymia}
Jungo, A., et~al: pymia: A python package for data handling and evaluation in
  deep learning-based medical image analysis. Computer methods and programs in
  biomedicine \textbf{198}, 105796 (2021)

\bibitem[{Kofler et~al.(2021)Kofler, Ezhov, Isensee, Balsiger, Berger, Koerner,
  Paetzold, Li, Shit, McKinley, Bakas, Zimmer, Ankerst, Kirschke, Wiestler, and
  Menze}]{kofler2021using}
Kofler, F., Ezhov, I., Isensee, F., Balsiger, F., Berger, C., Koerner, M.,
  Paetzold, J., Li, H., Shit, S., McKinley, R., Bakas, S., Zimmer, C., Ankerst,
  D., Kirschke, J., Wiestler, B., Menze, B.H.: Are we using appropriate
  segmentation metrics? identifying correlates of human expert perception for
  cnn training beyond rolling the dice coefficient (2021)

\bibitem[{Kofler et~al.(2022)Kofler, Wahle, Ezhov, Wagner, Al-Maskari, Gryska,
  Todorov, Bukas, Meissen, Peng et~al.}]{kofler2022approaching}
Kofler, F., Wahle, J., Ezhov, I., Wagner, S., Al-Maskari, R., Gryska, E.,
  Todorov, M., Bukas, C., Meissen, F., Peng, T., et~al.: Approaching peak
  ground truth. arXiv preprint arXiv:2301.00243  (2022)

\bibitem[{Lin et~al.(2017)Lin, Goyal, Girshick, He, and
  Doll{\'a}r}]{lin2017focal}
Lin, T.Y., Goyal, P., Girshick, R., He, K., Doll{\'a}r, P.: Focal loss for
  dense object detection. In: Proceedings of the IEEE international conference
  on computer vision, pp. 2980--2988 (2017)

\bibitem[{Ma et~al.(2021)Ma, Chen, Ng, Huang, Li, Li, Yang, and
  Martel}]{ma2021loss}
Ma, J., Chen, J., Ng, M., Huang, R., Li, Y., Li, C., Yang, X., Martel, A.L.:
  Loss odyssey in medical image segmentation. Medical Image Analysis p. 102035
  (2021)

\bibitem[{Milletari et~al.(2016)Milletari, Navab, and Ahmadi}]{milletari2016v}
Milletari, F., Navab, N., Ahmadi, S.A.: V-net: Fully convolutional neural
  networks for volumetric medical image segmentation. In: 2016 fourth
  international conference on 3D vision (3DV), pp. 565--571, IEEE (2016)

\bibitem[{Misra(2019)}]{misra2019mish}
Misra, D.: Mish: A self regularized non-monotonic neural activation function.
  arXiv preprint arXiv:1908.08681  (2019)

\bibitem[{Pan et~al.(2019)Pan, Schoppe, Parra-Damas, Cai, Todorov, Gondi, von
  Neubeck, B{\"o}{\u{g}}{\"u}rc{\"u}-Seidel, Seidel, Sleiman
  et~al.}]{pan2019deep}
Pan, C., Schoppe, O., Parra-Damas, A., Cai, R., Todorov, M.I., Gondi, G., von
  Neubeck, B., B{\"o}{\u{g}}{\"u}rc{\"u}-Seidel, N., Seidel, S., Sleiman, K.,
  et~al.: Deep learning reveals cancer metastasis and therapeutic antibody
  targeting in the entire body. Cell \textbf{179}(7), 1661--1676 (2019)

\bibitem[{Rahman and Wang(2016)}]{rahman2016optimizing}
Rahman, M.A., Wang, Y.: Optimizing intersection-over-union in deep neural
  networks for image segmentation. In: International symposium on visual
  computing, pp. 234--244, Springer (2016)

\bibitem[{Ronneberger et~al.(2015)Ronneberger, Fischer, and
  Brox}]{ronneberger2015u}
Ronneberger, O., Fischer, P., Brox, T.: U-net: Convolutional networks for
  biomedical image segmentation. In: International Conference on Medical image
  computing and computer-assisted intervention, pp. 234--241, Springer (2015)

\bibitem[{Salehi et~al.(2017)Salehi, Erdogmus, and
  Gholipour}]{salehi2017tversky}
Salehi, S.S.M., Erdogmus, D., Gholipour, A.: Tversky loss function for image
  segmentation using 3d fully convolutional deep networks. In: International
  workshop on machine learning in medical imaging, pp. 379--387, Springer
  (2017)

\bibitem[{Shirokikh et~al.(2020)Shirokikh, Shevtsov, Kurmukov, Dalechina,
  Krivov, Kostjuchenko, Golanov, and Belyaev}]{shirokikh2020universal}
Shirokikh, B., Shevtsov, A., Kurmukov, A., Dalechina, A., Krivov, E.,
  Kostjuchenko, V., Golanov, A., Belyaev, M.: Universal loss reweighting to
  balance lesion size inequality in 3d medical image segmentation. In:
  International Conference on Medical Image Computing and Computer-Assisted
  Intervention, pp. 523--532, Springer (2020)

\bibitem[{Silversmith(2021)}]{william_silversmith_2021_5535251}
Silversmith, W.: {seung-lab/connected-components-3d: Zenodo Release v1}. Zenodo
   (Sep 2021), \doi{10.5281/zenodo.5535251},
  \urlprefix\url{https://doi.org/10.5281/zenodo.5535251}

\bibitem[{Sirinukunwattana et~al.(2015)Sirinukunwattana, Snead, and
  Rajpoot}]{sirinukunwattana2015stochastic}
Sirinukunwattana, K., Snead, D.R., Rajpoot, N.M.: A stochastic polygons model
  for glandular structures in colon histology images. IEEE transactions on
  medical imaging \textbf{34}(11), 2366--2378 (2015)

\bibitem[{Sudre et~al.(2017)Sudre, Li, Vercauteren, Ourselin, and
  Cardoso}]{sudre2017generalised}
Sudre, C.H., Li, W., Vercauteren, T., Ourselin, S., Cardoso, M.J.: Generalised
  dice overlap as a deep learning loss function for highly unbalanced
  segmentations. In: Deep learning in medical image analysis and multimodal
  learning for clinical decision support, pp. 240--248, Springer (2017)

\bibitem[{Wright and Demeure(2021)}]{wright2021ranger21}
Wright, L., Demeure, N.: Ranger21: a synergistic deep learning optimizer. arXiv
  preprint arXiv:2106.13731  (2021)

\bibitem[{Zhang et~al.(2021)Zhang, Zhang, Li, Sweeney, Spincemaille, Nguyen,
  Gauthier, and Wang}]{zhang2021all}
Zhang, H., Zhang, J., Li, C., Sweeney, E.M., Spincemaille, P., Nguyen, T.D.,
  Gauthier, S.A., Wang, Y.: All-net: Anatomical information lesion-wise loss
  function integrated into neural network for multiple sclerosis lesion
  segmentation. NeuroImage: Clinical p. 102854 (2021)

\bibitem[{Zhao et~al.(2020)Zhao, Todorov, Cai, Rami, Steinke, Kemter, Mai,
  Rong, Warmer, Stanic et~al.}]{zhao2020cellular}
Zhao, S., Todorov, M.I., Cai, R., Rami, A.M., Steinke, H., Kemter, E., Mai, H.,
  Rong, Z., Warmer, M., Stanic, K., et~al.: Cellular and molecular probing of
  intact human organs. Cell \textbf{180}(4), 796--812 (2020)

\bibitem[{Zhu et~al.(2019)Zhu, Huang, Zeng, Chen, Liu, Qian, Du, Fan, and
  Xie}]{zhu2019anatomynet}
Zhu, W., Huang, Y., Zeng, L., Chen, X., Liu, Y., Qian, Z., Du, N., Fan, W.,
  Xie, X.: Anatomynet: deep learning for fast and fully automated whole-volume
  segmentation of head and neck anatomy. Medical physics \textbf{46}(2),
  576--589 (2019)

\end{thebibliography}

\end{document}